**Title:**

# Simulating realistic short tandem repeat capillary electrophoretic signal using a generative adversarial network

**Authors:**

Duncan Taylor[1,2] (Duncan.Taylor@sa.gov.au), Melissa A. Humphries[3] (melissa.humphries@adelaide.edu.au)

1. Forensic Science SA, GPO Box 2790, Adelaide, SA 5001, Australia
2. College of Science & Engineering, Flinders University, GPO Box 2100, Adelaide, SA 5001, Australia
3. School of Computer and Mathematical Sciences, University of Adelaide, North Terrace Campus, Adelaide, SA, Australia 5005

**Highlights:**

- We present a pipeline to produce fully simulated, realistic DNA profiles
- Generative adversarial networks (GAN) are used to simulate time-series like outputs
- A novel construction of ANNs for electrophoretic inputs
- Biologically informed GAN increases relevance and accuracy
- A modification of the pix2pix architecture for non-square, time-series like, multivariate data.
- Proficiency over small data (1078 profiles)

**Abstract:**

DNA profiles are made up from multiple series of electrophoretic signal measuring fluorescence over time. Typically, human DNA analysts 'read' DNA profiles using their experience to distinguish instrument noise, artefactual signal, and signal corresponding to DNA fragments of interest. Recent work has developed an artificial neural network (ANN) to carry out the task of classifying fluorescence types into categories in DNA profile electrophoretic signal. But the creation of the necessarily large amount of labelled training data for the ANN is time consuming and expensive, and a limiting factor in the ability to robustly train the ANN. If realistic, pre-labelled, training data could be simulated then this would remove the barrier to training an ANN with high efficacy. Here we develop a generative adversarial network (GAN), modified from the pix2pix GAN to achieve this task. With 1078 DNA profiles we train the GAN and achieve the ability to simulate DNA profile information, and then use the generator from the GAN as a 'realism filter' that applies the noise and artefact elements exhibited in typical electrophoretic signal.

## 1.0 - Introduction

The process of classifying the fluorescence in the DNA profile is predominantly done by, commonly two, human readers in forensic biology laboratories. This process is time consuming and produces hard (as opposed to probabilistic) classifications of features into multiple categories. Artificial intelligence (AI) can be used to replace at least one of the human readers (Taylor, 2022; Taylor, Harrison, & Powers, 2017; Taylor, Kitselaar, & Powers, 2019; Taylor & Powers, 2016) and provides probabilistic classifications that can potentially increase the accuracy of DNA profile analysis (Taylor & Buckleton, 2023). However, the efficacy of the AI systems has been limited due to a lack of large amounts of labelled data. This paper presents a method for simulating the highly realistic DNA profiles required to train effective AI classification systems using biologically informed generative adversarial networks (GANs).

### 1.1 – Classifying DNA profiles

Short tandem repeat (STR) DNA profiles are produced by passing fluorescently tagged amplicons through a gel-filled capillary (separating the fragments according to their size) and past a laser and detector (Butler, 2009). The greater the amount of starting DNA, the more amplified DNA fragments will be present (each with an attached fluorophore) and the greater the detected fluorescent signal will be. The strength of these detected signals is represented in electropherograms, displaying fluorescence over time, measured in relative fluorescence units (rfu). The electropherogram appears as a time series-like output with varying sized peaks representative of the features of the profile. Each point in the series is referred to as a 'scan point' in the raw signal, but are processed to represent the base pairs making up the fragment size in the final DNA profile). Modern DNA profiling systems utilise multiple fluorophores, which fluoresce at different wavelengths, and so a modern DNA profile will display the fluorescent signal at multiple wavelengths over time.

As with all real-world processes the raw electrophoretic signal is not a smooth, clean and clear signal. The capillary electrophoresis instrument generates a level of baseline noise, there are incomplete separation of detection of light at different wavelength, there are unincorporated DNA profiling reaction components that lead to non-specific fluorescence ‘mounds’, the baseline of the instrument can drift over the course of the run, and inclusions in the capillary or gel can cause fluorescence ‘spikes’ in the signal. The resulting DNA profile is a combination of baseline noise, artefactual signal and the signal of interest. It is common practice is for two independent analysts to interpret the signal of interest in DNA profiles (in a process called ‘reading’ the DNA profile), compare results, resolve differences, and ultimately produce a processed profile that can be used in downstream DNA profile interpretation and evaluation.

Recent work has developed an artificial neural network (ANN) that can classify the fluorescence in the DNA profile into one of several categories including alleles, artefacts and baseline measures (Taylor, 2022; Taylor, Harrison, et al., 2017; Taylor et al., 2019; Taylor & Powers, 2016). This ANN has already been used to replace one of the two human readers in some forensic biology laboratories (Volgin, Taylor, Bright, & Lin, 2021). Apart from a clear time-saving benefit, the ANN also produces probabilistic classifications that hold more information than the binary classifications made by human readers. Additional work has shown that the probabilities associated with using the ANN to classify peaks can be used directly in DNA profile analysis (Taylor & Buckleton, 2023), removing the need for human readers at all (also see (Taylor & Abarno, 2023) which utilises this approach). However, replacing both human readers with the ANN would require an extremely high performing system.

It is well-known that the performance of ANNs is closely tied to the amount of training data available. As with many supervised learning tasks, the ANN developed in (Taylor, 2022) relies on large amounts of manually labelled training data, which can take extensive time to create. It is infeasible to recruit humans to manually label the amount of data required due to those time constraints. Taylor (2022) was able to augment their training set by modifying existing labelled data, but it could only create one low-level equivalent of each real profile (hence, at most, only doubling the size of the data). As such, the modified data process had limited effect and would not generate new combinations of peaks. Therefore, defining way of automatically labelling training data, or simulating highly realistic training data will provide enormous benefits to training an ANN.

### 1.2 – Biologically informed Generative Adversarial Networks (GAN)

One potential avenue that could simulate highly realistic electrophoretic signal is the use of generative adversarial networks (GAN) (Goodfellow et al., 2014). GANs are a system of competing ANNs, in which one ANN generates data designed to look as realistic as possible (the generator) and the other ANN discriminates between the generated data and real examples (the discriminator). As the GAN training progresses, the generator gets better at producing realistic data, and the discriminator gets better at discriminating between real and generated data, until the two systems have reached an equilibrium. At this point the generator is (ideally) generating highly realistic data.

Classically, GANs are set up to simulate random, realistic data. For example, in the context of DNA profiles, the generator may learn to generate realistic looking signal from supplied random input. In this case, the output is also random and there is no control over the generated features of the signal. There are two practical limitations to using purely random input:

1) Without defining where allelic signal is expected, pre-labelling of the simulated data as allele or artifact becomes impossible, and
2) data with specific properties cannot be simulated to fill training data gaps that exist in the current set.

But the deepest limitation, keeping in mind that a lack of large amounts of data is driving this exploration, is that the GAN from a purely random input is unlikely to learn the nuances of the biological models that make position and height of peaks realistic (i.e., the level of peak height variability expected for peaks at different heights, the pairing of peaks in complex mixed profiles, the generation of stutter products during PCR). Meaning the output may loosely look real but be biologically invalid.

For this reason, the system of DNA profile simulation we propose uses biological models to simulate realistic information about the number, size and height of peaks expected in a DNA profile, and then will use a trained GANs to essentially apply a 'realism filter' to the simulated data. This combination of biological and generative models provides an ability to create unlimited, pre-labelled training data that looks highly realistic, and can be used to train a DNA profile classification ANN.

## 2.0 - Method

All calculations were carried out using R V4.2.3 ("R Core Team. R: A language and environment for statistical computing," 2013) using packages tensorflow V2.11.0 (Abadi et al., 2015), Keras V2.11.1 (Chollet, 2015), and simDNAmixtures V1.0.1 (Kruijver, Kelly, Bright, & Buckleton, 2023).

### 2.1 – Input data

A total of 1078 GlobalFiler™ DNA profiles were used in the training of the GAN. Of the 1078 profiles, 669 were obtained from the online PROVEDIt dataset (Alfonse, Garrett, Lun, Duffy, & Grgicak, 2017) (taken from the GlobalFiler™ 29 cycle PROVEDIt, 2-5 Person profiles run on a ABI3500 with injection time of 5sec). The remaining 409 profiles were taken from the GlobalFiler™ profile data run on ABI3500 used as training data by Taylor (Taylor, 2022) and de-identified, mixed casework profiles.

The DNA profiles ranged from being blank, with only baseline information present, to complex 5-person mixtures. From the PROVEDIt dataset the count of 2 to 5 person mixtures was 167, 141, 149, and 137 respectively (with the remaining 75 PROVEDIt samples being blanks). Peak heights ranged up to 32750 rfu (30 000rfu is typically considered the saturation level of the laboratory instrument used to capture EPG fluorescence and 33 000rfu is the upper capture limit).

Ladders were removed from the input data, however negative and positive controls were retained. HID files were converted to csv format, which held fluorescence at each scan point in each dye lane across the whole profile (approximately 10 000 scan points). Only scan points between 4000 and 9000 were used in training and generation as prior to 4000 the electrophoretic signal associated with primer flare and is not suitable for training, and beyond 9000 no peaks are expected. GlobalFiler™ produces peaks labelled with 6 different fluorophores so the data for each profile is represented by a [6 x 5000] matrix of values. Values for rfu range between -30 000 and 30 000. Prior to use in any ANN, the mode of the fluorescence is then subtracted from each profile (acting as a simple baselining function, without changing the shape and patterns in the signal), and they are scaled down by a factor. Different scaling factors were trialled (data not shown) and a scaling factor of 100 was found to work best with both generator and discriminator.

The DNA profiles from the (Taylor, 2022) study possessed manually labelled fluorescence category information (as it was used in the training of an ANN to classify areas of electropherograms).

## 2.2 – Choice of GAN

A common input for the generator in GANs is a series of random numbers. The generator then learns to turn these random numbers into the desired output. Isola et al (Isola, Zhu, Zhou, & Efros, 2018) developed a GAN system called pix2pix that provides an image as the input to the generator, and the task for the generator was to convert that image to another image, meeting both the requirements of generating a realistic looking image in the target style, but also retaining the features of the original input. An example use for this type of task would be to take a rough line drawing of an object (such as a building, shoe, or face) and then convert it to a realistic image of that drawn object. The generated image would look like the real object, but retain the shape, size and style of the original sketch. For DNA profile generation the pix2pix GAN allows the flexibility to incorporate a biologically informed "sketch" of the DNA profile, making this an idea choice of GAN architecture. The pix2pix takes an idealised DNA profile (i.e., baseline of 0 and peaks represented by gaussian distributions) and learns to convert them to realistic counterparts.

There is a pix2pix package available for python (https://github.com/junyanz/pytorch-CycleGAN-and-pix2pix), however this was not suitable for the electropherogram simulation application due to the fact that the input was not a square image (required by the code), but rather effectively six time series with 5000 scan points. Also, electrophoretic input possesses specific patterns of dependence across scan points within a dye lane and at a particular scan point between dye lanes, but not all around a central point such as in an image. Therefore, while the general architecture of GAN was based on pix2pix, a novel construction of ANNs were innovated for electrophoretic inputs, and the generator loss function was modified from the original published construction.

### 2.3 – Preparing generator input

For training to occur the pix2pix generator requires paired input images, one in the style of a typical input the generator will be provided, and the other being its real equivalent. These paired images are created from real starting images. The following steps were undertaken to create a biologically driven, idealised version of a profile:

1. A baseline trend was determined by modelling the real profile with a lowess line using a smoother span of 0.05. This ensured a clean estimate of the trend, uninfluenced by the peaks. An example of the lowess line seen in Figure 1b.
2. The lowess trend was subtracted from the real DNA profile to create a de-trended profile that no longer possesses baseline drift but does possess baseline noise.
3. Using the de-trended profile, peaks were detected using the method of Woldegebriel (Woldegebriel, Asten, Kloosterman, & Vivó-Truyols, 2017), with two modifications. First, the algorithm weighed up the propositions that a peak was in the central 3 scan points of the window vs in any other scan point of the window, or no peak being present. Second, the context window first had the mode of the data within the window subtracted. This overcame some of the shortcomings of the lowess line (which did not always follow baseline drift completely, particularly when the drift caused relatively ‘peaky’ mounds). The cut-off for detecting a peak was that the probability for the central 3 scans possessing a peak centre was greater than $10^{10}$.
4. The original DNA profile was passed through the ANN from Taylor (2022), and any peak centres detected in point 3 that had been classified as baseline or pull-up (according to the category with the maximum probability assigned by the ANN) were removed. If the input DNA profiles already had manually assigned labels then these were used rather than generating new assignments.
5. The idealised profile was drawn with a baseline of 0 and peaks (with heights matching those in the de-trended real profile) drawn as normal distributions with a standard deviation of 4.

Figure 1 shows the various stages of profile smoothing.

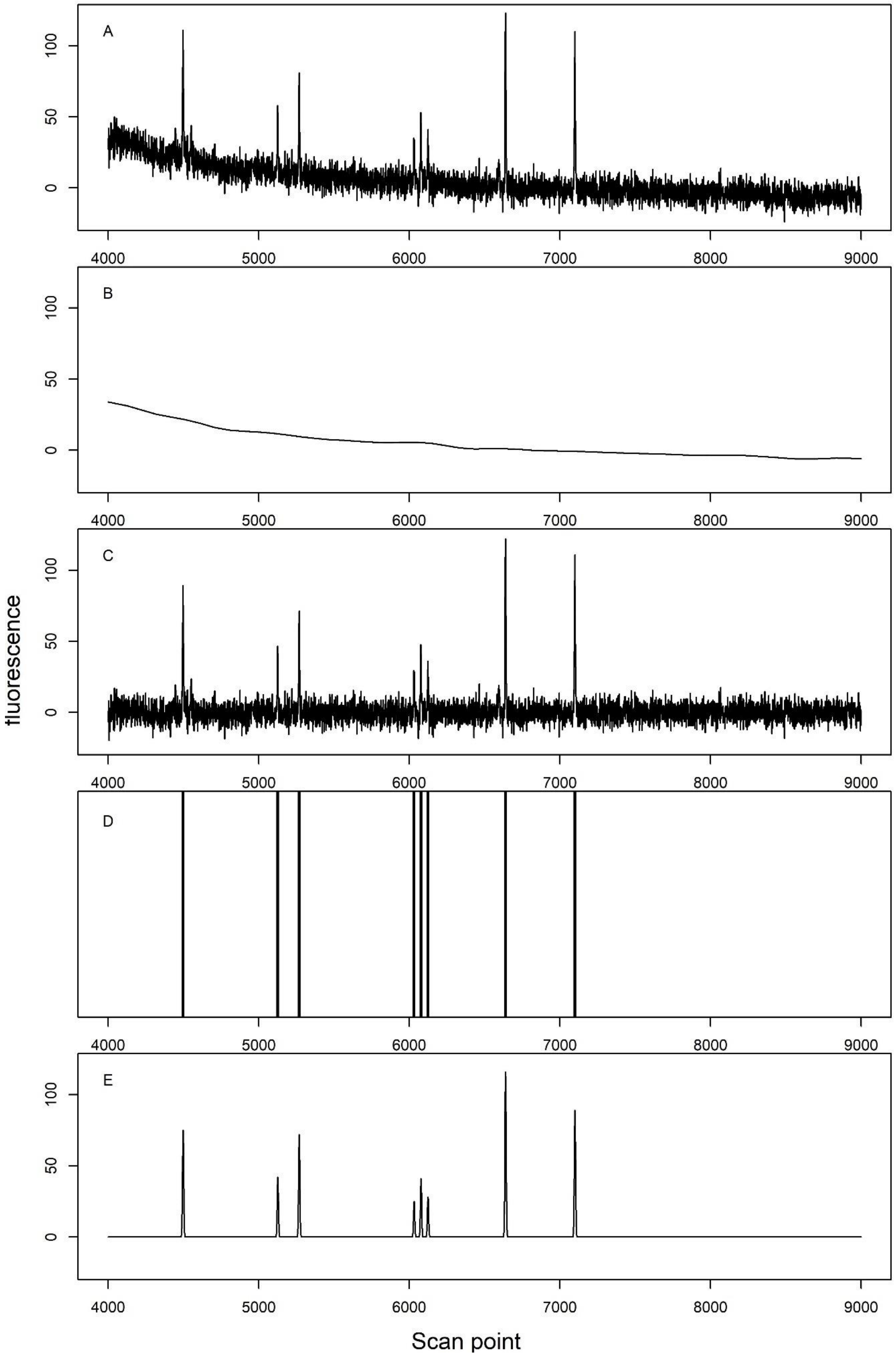

*Figure 1: Stages of creating a smoothed DNA profile. Panel A shows the original real DNA profile, panel B shows the lowess line used to model trend/baseline, panel C shows the real*

*profile with the lowess baseline subtracted (de-trended profile), panel D shows the results of the peak detection algorithm, and panel E shows the resulting, biologically driven, idealised DNA profile.*

This process was carried out for 1078 profiles to produce paired input and output profiles for use by the generator.

## 2.4 – The generator architecture

The generator in the original pix2pix GAN was an implementation of the U-Net convolution network designed by Ronnenberger et al for image segmentation (Ronneberger, Fischer, & Brox, 2015). The design of the U-Net is well suited for images, but not optimised for use on electrophoretic data. The architecture adapted for use on electrophoretic data, which was still based on the U-Net architecture, is shown in Figure 2:

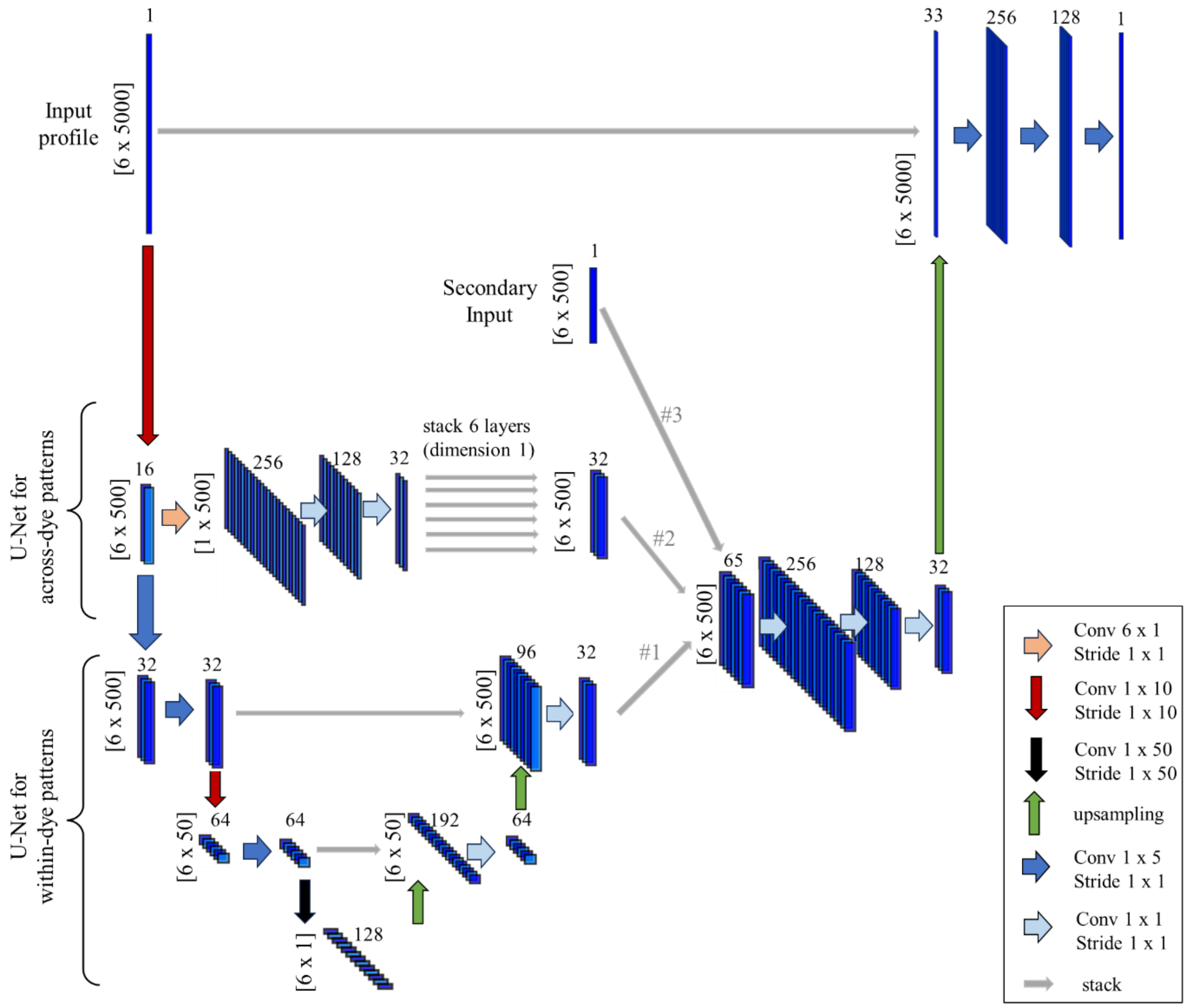

*Figure 2: Architecture of the generator ANN used in the GAN*

The architecture in Figure 2 has several features to note:

- After an initial downsizing of the input, there is a U-Net structure that uses convolutional layers with filter sizes [1 x N] that specifically target the within-dye patterns in the profile input.
- There is a secondary U-Net structure that uses a convolution layer with filter size of [6 x 1] that specifically target the across-dye patterns in the profile input.
- The within-dye layers and the across-dye layers come together for the final upsizing to the output. The across-dye features are of dimension [1 x 500 x 32] and apply, but act differently, to each dye. Across-dye layers are therefore duplicated and stacked to produce dimension [6 x 500 x 32] - allowing stacking with the within-dye layer of size [6 x 5 x 32].
- There is a secondary input of size [6 x 500 x 1] that stacks simultaneously with both the within- and across-dye layer stacks. This secondary input allows a structured or randomised element to be introduced into the GAN, both of which were trialled in this work.
- After each convolutional layer a ReLu activation function was applied, followed by batch normalisation. The only exception to this was the final output layer, which utilised a leaky ReLu activation and no batch normalisation. Also, in the final convolutional layers 2D spatial dropout was applied with a rate of 0.25.
- The ANN shown in Figure 2 has 1,433,361 parameters.

## 2.5 – The discriminator architecture

The discriminator architecture from the pix2pix implementation was used, with modified filter sizes applicable to the [6 x 5000] input. The structure of the discriminator is shown in Figure 3.

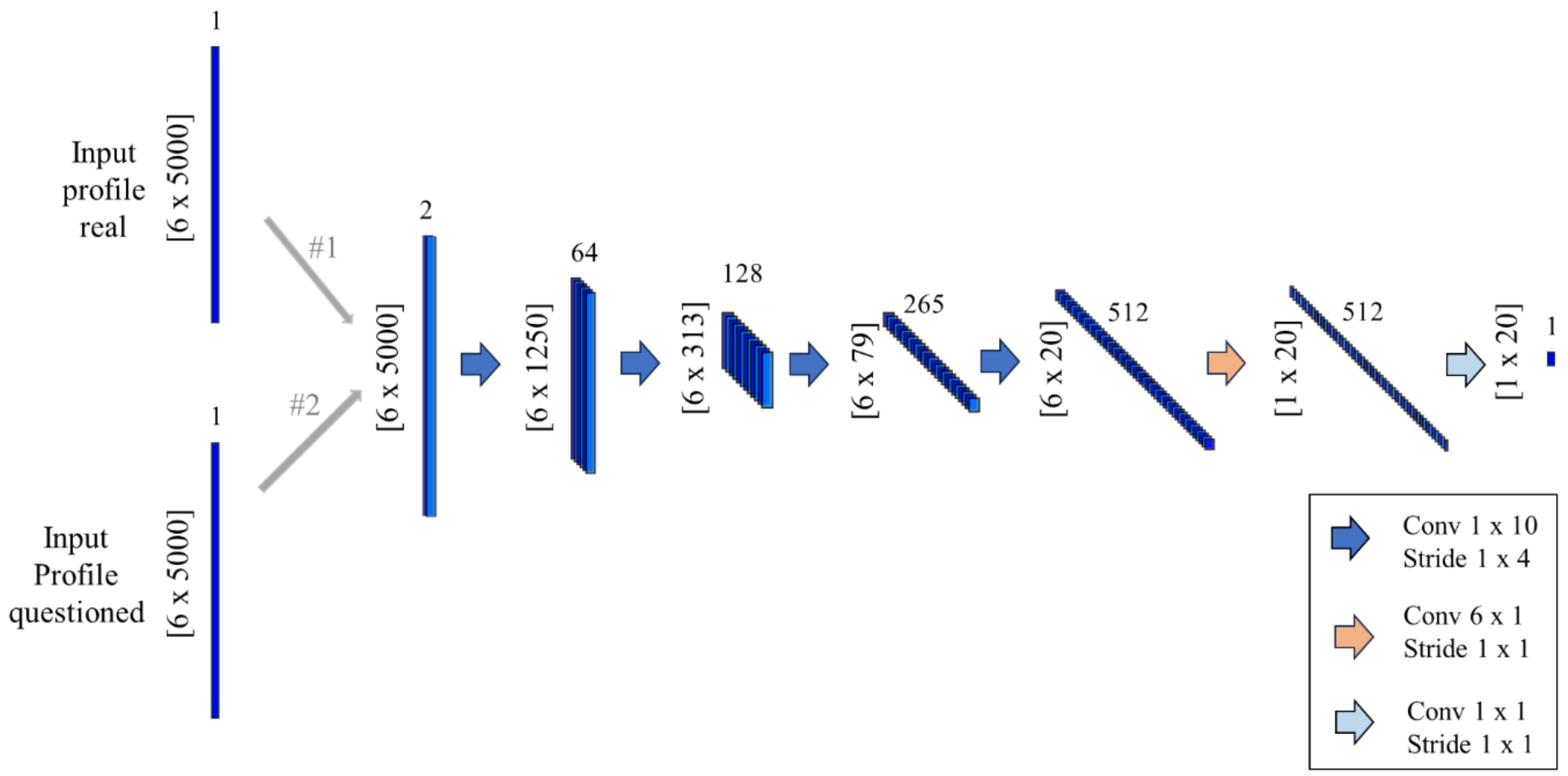

*Figure 3: Architecture of the discriminator ANN used in the GAN*

For the ANN in Figure 3:

- All but the final layer has a leaky ReLu activation function. The final output has a sigmoid activation.
- After each convolutional layer batch normalisation was applied.
- The final output is a [1 x 20] array, which represents 20 segments of the learned profile features. This type of multi-output in a discriminator is based on the PatchGAN architecture (proposed by (Wand, 2016)).
- The ANN shown in Figure 3 has 3,302,081 parameters.

<u>2.6 – Training the GAN</u>

In general, a GAN works by using a pair of competing ANNs: a generator and a discriminator. The generator has the function of producing an output that adheres to a specific goal, usually to produce highly realistic images. The discriminator has the function of distinguishing between the real and fake images (i.e., comparing a known, true image to the generated image). The discriminator performance is judged by how well it distinguishes between the real and generated images. The generator performance is judged by how well it can produce images that fool the discriminator. The input of a classic generator is an array of random numbers, which it learns to translate into an image. In the pix2pix GAN the performance of the generator is judged on two criteria, the ability to fool the discriminator, but also the ability for the output to match features of the target.

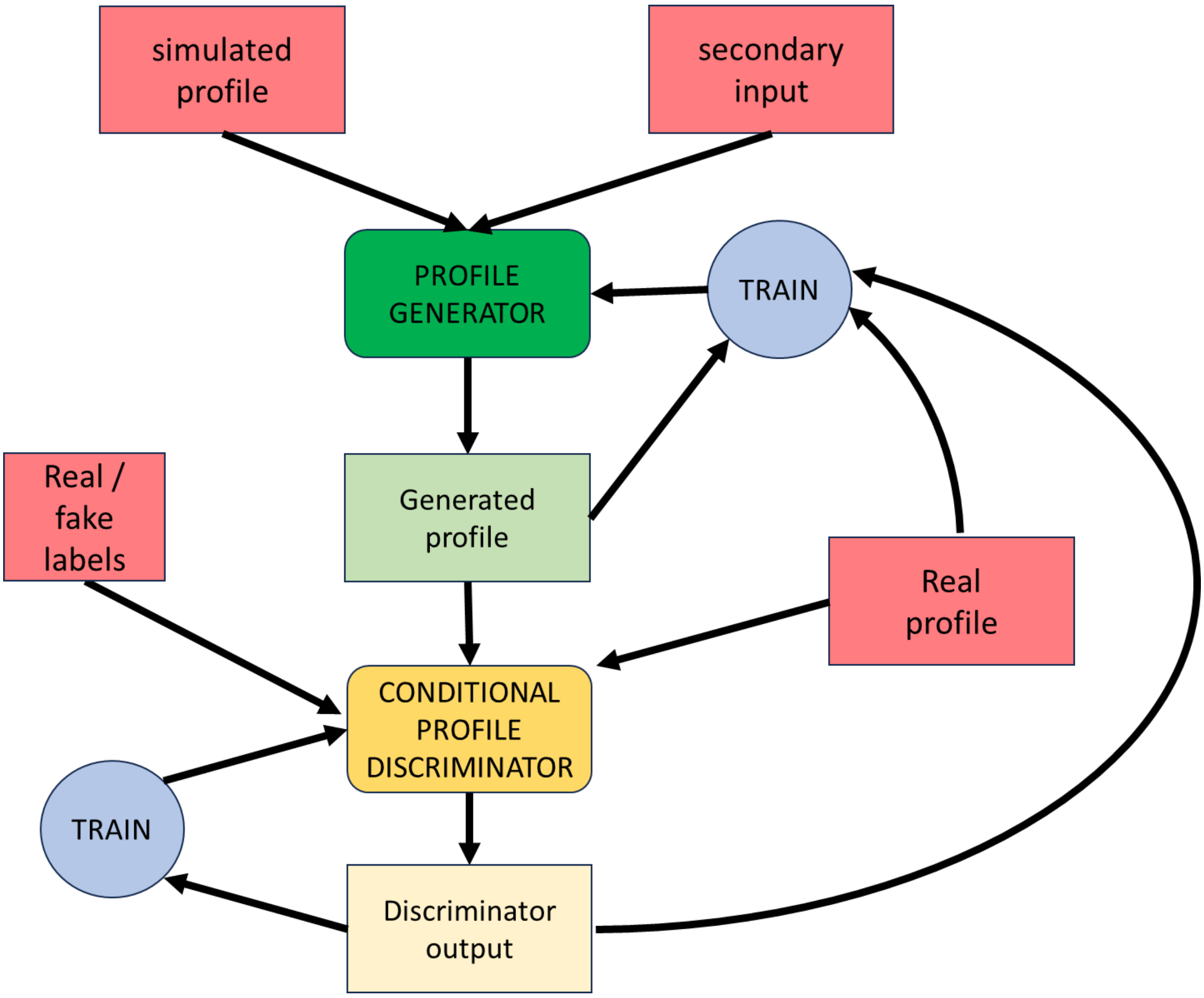

*Figure 4: Architecture of the pix2pix GAN variant used to simulate DNA profiles. pink coloured boxes represent inputs, green boxes represent the generator (dark green) and the generator output (light green). Yellow boxes represent the discriminator (dark yellow) and discriminator output (light yellow). Blue circles represent training steps.*

The architecture of the GAN used to simulate electropherograms is shown in Figure 4. The profile generator takes two inputs, the biologically informed, idealised profile and a secondary input (which we trial as either being random or structured values). It uses these to generate a profile. The generated profile is then passed into the discriminator, along with the real profile and labels that designate the profiles as real or fake. This type of discriminator is known as a conditional discriminator. It is conditioned on the information in the starting, smoothed profile and is asked to determine whether a second profile (either generated from that smooth profile, or the original real starting profile) is real or fake. The generator then takes the output of the discriminator, the real profile and the initial generated profile. This cycle continues to until both ANNs have reached converged to a steady state of performance, based here on a loss function.

The loss function for the discriminator is the sigmoid cross entropy between the labels for the real profiles and their labels, plus the sigmoid cross entropy between the labels for the generated profiles and their labels. The labels for real profiles are 1 and for generated profiles are 0, however, to assist in avoiding mode collapse, we employ 'soft labels' by adding random values drawn from U[0, 0.05] to each label.

The loss function for the generator is the sigmoid cross entropy between the labels for the real and generated profiles, and the labels for the generator are compared to their opposing extrema. The discriminator loss is minimised by yielding outputs to real profiles close to 1, and outputs to generated profiles close to 0. The generator loss is minimised by the discriminator yielding outputs for generated profiles close to 1. Soft labels are also used for this task. The total generator loss is calculated by the weighted sum of the mean of the absolute difference between the generated output and the target image. We found that for the task of simulating electrophoretic data the weighting between the two components needed to be equal, contrary to the original publication which found that the second component (closeness to the target image) needed to be weighted 100-fold higher than the first component (ability to fool the discriminator).

For both generator and discriminator ANNs the Adam Optimiser (Kingma & Ba, 2017) was used with learning rate set to $5\times10^{-5}$ , beta1 set to 0.5 and beta2 set to 0.999.

GANs can be subject to mode collapse - where the generator learns to create one image that can fool the discriminator, and never varies from it. A traditional GAN uses randomly generated inputs, allowing for multiple variants of the desired output to be created, to aid in prevention of mode collapse. In the original pix2pix GAN the authors found that, unlike traditional GANs, random input was not required (along with the image input already being provided) and was ignored as the GAN learned. We trialled a secondary input with our version of the pix2pix GAN. Two variants of the secondary input were trialled, one being [6 x 500] random values drawn from a U[0, 1], and the second being a structured input of values from 0 to 1 in increments of 0.002 to produce a [1 x 500] array repeated (stacked) six times to produce a [6 x 500] array. The intuition behind the second variant was based on the fact that electropherogram will often have a gentle exponential decay pattern to baseline (Bright, Taylor, J.M., & Buckleton, 2013). This is a profile-wide feature, which requires learning at the deepest layer in the U-Net. Supplying the structured array as secondary input provide location information for the convolutional layers working on only a small part of the profile and may assist in the quality of generations.

To train the GAN, pre-training was conducted on the generator ANN and discriminator ANN separately so that they started their adversary with some knowledge on how to perform their tasks. For each ANN 100 epochs of the full 1078 training set were carried out, with a single batch of 1078. The generator ANN carried this out in two stages of 50 epochs. In the first stage sample weights were supplied (idealised profile plus one) and in the second stage no sample weights were supplied - ensuring that the peak information is upweighted in comparison to the baseline information. The GAN was run for a further 200 epochs with a batch size of 2 (leading to 539 batches). All training was conducted on an Intel® Xenon® with E3-1505M v5 CPU @2.80 GHz and 64 GB RAM with 64-bit Windows 10 Professional. Training (including the ANN pre-training and GAN training) took approximately 96 hours to complete.

## 2.7 – Simulating biologically informed random profiles

After the GAN training was complete the generator ANN could be used to create realistic DNA profiles from purely simulated data. The R package simDNAmixtures (Kruijver et al., 2023) was used to simulate DNA profile information. It was set up to simulate profiles by:

- drawing alleles from the Australia Caucasian population (Taylor, Bright, McGovern, Neville, & Grover, 2017),
- drawing template and degradation values from within user-defined bounds,
- applying back, forward, half-back and double back stutters to allelic peaks based on their expected ratios,
- applying peak height stochastic variability to all peaks based on their fluorescence type, and
- applying inter-locus imbalance.

Expected stutter ratios and peak and locus balances were based on in-house validation of the GlobalFiler™ profiling kit at the lead author's laboratory.

Once simulated the profiles were supplemented in three ways:

- Amelogenin peaks were added by randomly choosing a sex for each contributor and using their simulated DNA amounts to generate X and potentially Y peaks.
- Peaks were added for each internal lane standard (ILS) peak for the Thermo Fischer Scientific GeneScan™ 600 LIZ™ dye Size Standard v2.0.

- A column was added that had a scan point for each peak, which was carried out by multiplying the base pairs by 11.2 and adding 3500. These values were obtained by graphing ILS base pairs vs scans for 20 samples.

Given this construction the simulated profile data was then converted to a smoothed idealised profile by setting a baseline of zero and adding modelling peaks with a normal distribution with a mean equal to the scan point of the peak centres and a standard deviation of 4. The fluorescence for each peak was scaled by the corresponding simulated peak height. This smooth electropherogram was then passed through the generator ANN, which acted as a 'realism filter'. At this point a realistic electropherogram was available that had been completed simulated.

**3.0 - Results**

Using either a structured or secondary input yielded approximately equivalent results. We do not show the results here, however supplying the structured input lead to the convergence of the GAN in approximately half the epochs was required by the randomised secondary input. However, once trained, the generator with the randomised input has other advantages (which we discuss later) that means that it was ultimately chosen to proceed with. All results shown below are for the generator that uses a random input.

<u>3.1 – Results of training GAN</u>

Figure 5 shows the average loss, across the entire dataset, of the GAN training across 200 epochs. The initial very high performance of the discriminator is due to the pre-training (during which the generator output is static) and is the state of the ANN prior to the GAN training. There is an initial decrease followed by an increase in loss for generator. These changes are due to the changing performance of the discriminator rather than a drift of the generated profile away from the real profile. The lower panel of Figure 5 shows the performance of the discriminator at each of the 10 epochs on the real and generated profiles. The results in Figure 5 show the discriminator performance (and the GAN in general) had converged by approximately 140 epochs in this dataset of 1078 profiles. After this point the loss for both discriminator and generator plateaued and the ability for the discriminator to identify real and simulated profiles remained approximately constant.

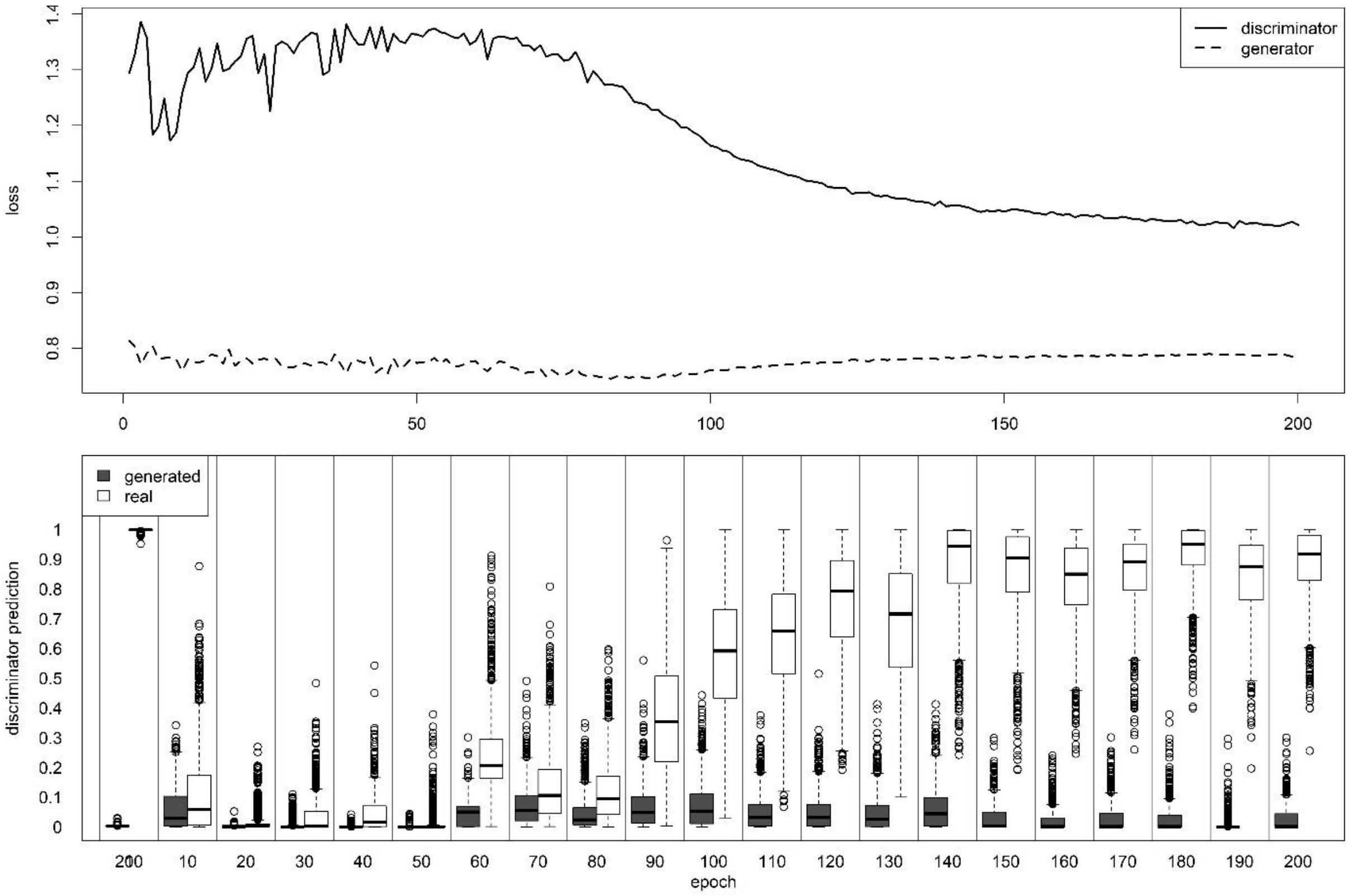

*Figure 5: A) loss for generator and discriminator during the GAN training and B) performance of the discriminator shown at every 10 epochs on generated (grey) and real (white) profiles.*

Figure 6 shows an example of the performance of the generator. In the generated profile, there has been generated instrument noise added to the smoothed, input profile. This is a random element of profile generation that can only be learned by the performance of the discriminator in the GAN (as opposed to some identifiable pattern, which can be learned by the generator). In other profiles (most noticeable in electropherograms with low intensity allelic peaks) another random element in real profiles is larger trends in baseline drift. There were instances of the generator adding these baseline drift events into profiles which did not originally have them or had different patterns of baseline drift. An example of this type of addition can be seen in Figure 6, which shows the real profile, the smoothed profile, and the generated profile. The real profile had a pattern of baseline drift that undulated over 100 scan points, whereas the generated profile was relatively flat across the same scan range but had a steep trend in the first 200 scans. There were, however, many instances of baseline drift in the original profiles that were also similarly replicated in the generated profiles, which suggests that there was some overfitting of the model to the dataset.

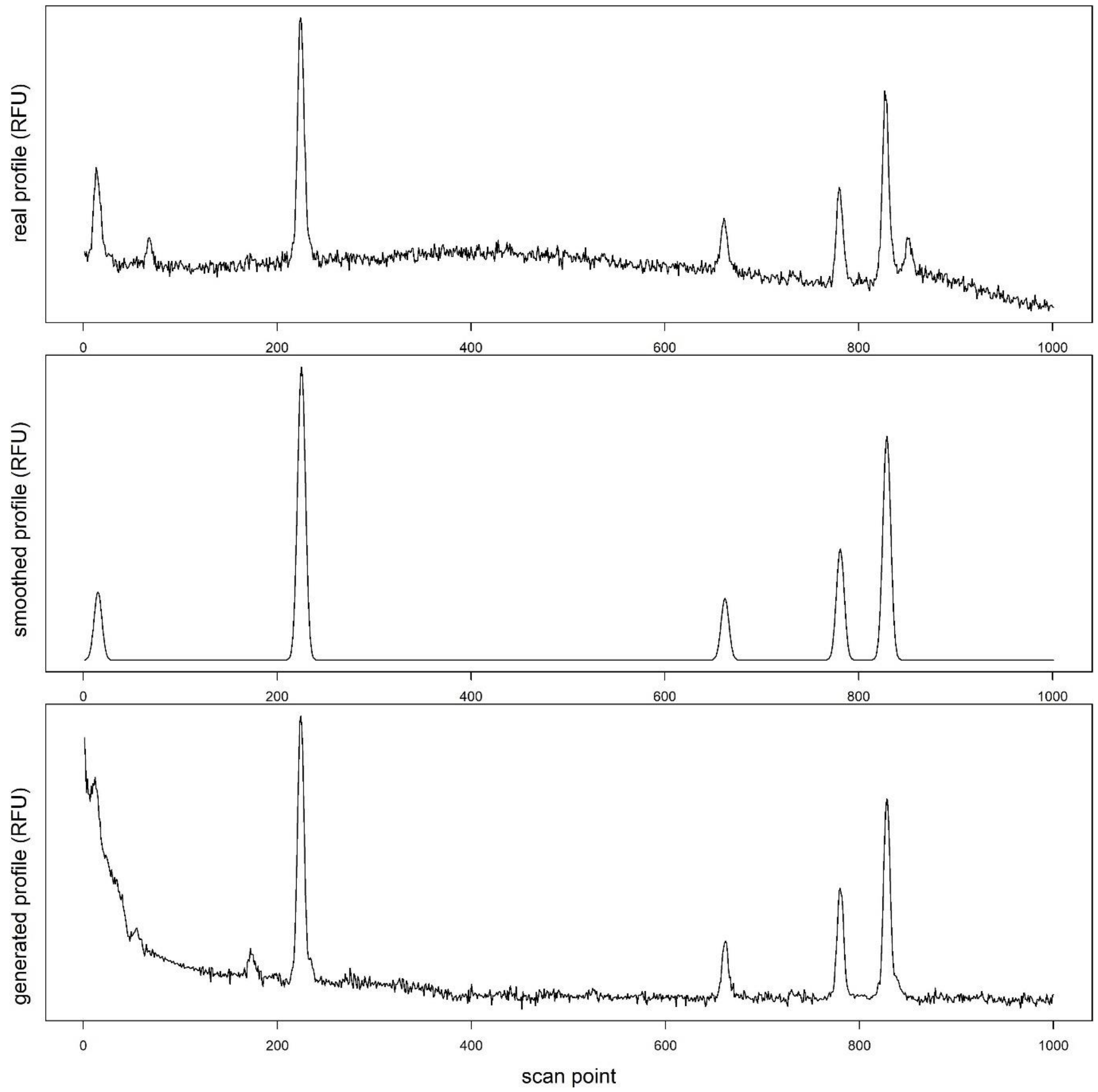

*Figure 6: The first 500 scan points of dye lane of a DNA profile used in the dataset with the real profile shown in the upper panel, the smoothed profile (being used as the generator input) in the middle panel and an example of a generated profile in the lower panel.*

As well as random profile features, there were also features learnt by the generator that are structured, but not present in the smoothed input profile. One such structure is the artefactual, ‘pull-up’ peak. Pull-up peaks occur due to the fluorescence detection for a specific dye (occurring within a specific window of wavelength) detecting fluorescence signal from another dye (see (Taylor & Powers, 2016)). On an electropherogram these appear as ether small peaks, or small troughs that align with intense peaks in other dye lanes. Figure 7 shows several instances of pull-up that were present in the original profile, not present in the smoothed input profile, but has been learnt by the generator (for example the dip in the centre of the lowest dye lane). Another example of structured data that was learn by the generator is the presence of stutter peaks. These occur as small peaks in earlier scan point to an intense ‘parent’ peak. In Figure 7 an instance of this can be seen in the third dye lane within the right-hand cluster of peaks. In the original profile there is an absence of the expected stutter peak (this is a phenomenon known as peak drop-out and occurs randomly for low intensity peaks). The smoothed profile therefore also does not have an observed peak at this position. The generator has learned the presence of these stutter peaks and in Figure 7 the generated profile has the stutter peak present.

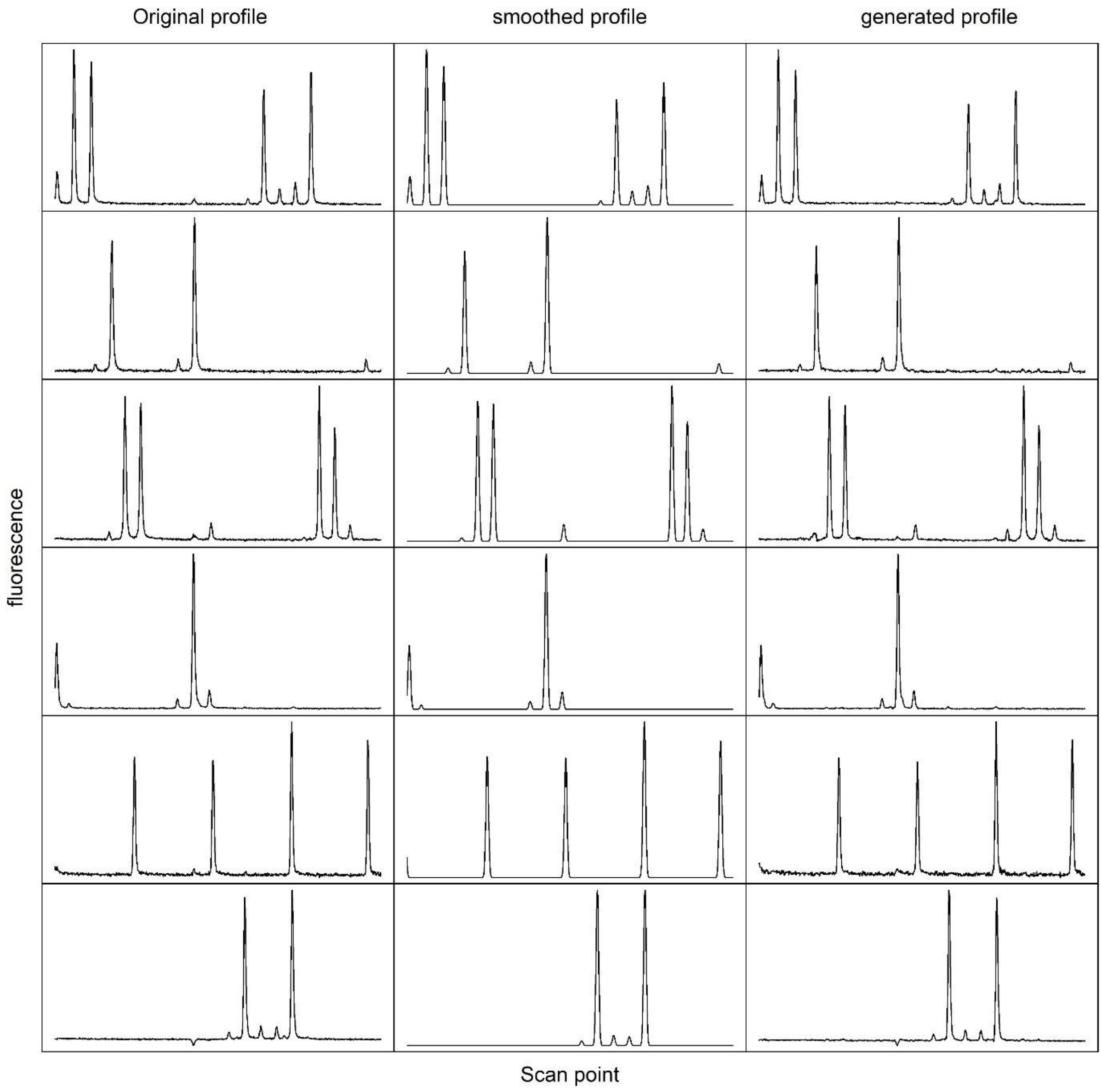

*Figure 7: Example of generator performance after 100 epochs. The left panel shows the original profile, the middle panel shows the smoothed profile that is used as input to the*

*generator and right panel shows the generator output. Only a small scan point range (4500 to 5500) is shown to demonstrate performance.*

In order to gauge the performance of the generator, a confusion matrix was produced by taking a real DNA profile and classifying the fluorescence using the profile classification ANN from (Taylor, 2022). The DNA profile was then smoothed to produce an ideal input, then passed through the generator to recreate a simulated DNA profile and reclassifying using the profile reading ANN again. The confusion matrix was created by comparing the original vs generated classifications (Figure 8). A confusion matrix was generated for each of five test profiles from (Taylor, Harrison, et al., 2017) which were chosen to range from weak to intense. In the most intense profiles (1 and 2) there were instances of all categories of fluorescence in the original profiles, and as the profiles became weaker the categories of fluorescence observed dropped away in order of their general expected fluorescent intensity until the set was reduced to only baseline and allele (Profile 5). The half-back stutters or the forward stutters categories of fluorescence are generally the weakest peaks in profiles (given their expected heights relative to their parents (see (Taylor et al., 2016) for an example of expected forward stutter peak heights relative to parent peaks).

For allele, back stutter, forward stutter, and half stutter fluorescence categories that were classified in the real profile, a peak will exist. Whether the same region is still classified in the same way for the generated profile relies on; the performance of the smoothing algorithms (de-trending, and peak detection), the generator ANN, and whether the generator had added structural information into the profile that was not present in the original profile (as was shown occurring for a back stutter peak in Figure 6).

The greatest test of the generator to learn structural features of a profile is the pull-up category. If a region of fluorescence is classified as pull-up in the original profile it will not correspond to any feature in the smoothed profile (i.e., a peak or trough). Therefore, to be classified as pull-up again in the generated profile, the generator will need to have recognised the pattern of fluorescence around that point and added a pull-up feature. In Figure 8, across profiles 1 to 3, the originally classified pull-up scan points were classified as pull-up again in the generated profile in 50% to 75% of instances. This is a high level of performance given the very subtle patterns of fluorescence that are classified as pull-up, and the limited dataset used to train the generator (and within that dataset the limited number of profiles that reached the intensity required for pull-up to be present).

**Profile 1 (most intense)**

Rows: classification in generated profile; columns: classification in real profile

| | A | B | F | H | P | S |
|---|---|---|---|---|---|---|
| A | 1169 | 251 | 0 | 3 | 27 | 0 |
| B | 59 | 25543 | 15 | 1 | 44 | 26 |
| F | 8 | 166 | 143 | 0 | 0 | 2 |
| H | 0 | 13 | 0 | 29 | 0 | 0 |
| P | 48 | 591 | 0 | 0 | 1236 | 10 |
| S | 13 | 77 | 0 | 0 | 0 | 526 |

**Profile 2**

Rows: classification in generated profile; columns: classification in real profile

| | A | B | F | H | P | S |
|---|---|---|---|---|---|---|
| A | 1210 | 196 | 0 | 0 | 14 | 0 |
| B | 12 | 26205 | 16 | 0 | 17 | 41 |
| F | 0 | 98 | 57 | 0 | 0 | 0 |
| H | 1 | 21 | 0 | 11 | 0 | 0 |
| P | 44 | 258 | 0 | 0 | 1249 | 2 |
| S | 0 | 86 | 0 | 0 | 0 | 462 |

**Profile 3**

Rows: classification in generated profile; columns: classification in real profile

| | A | B | P | S |
|---|---|---|---|---|
| A | 938 | 104 | 0 | 0 |
| B | 34 | 28459 | 0 | 14 |
| P | 0 | 93 | 72 | 0 |
| S | 9 | 59 | 0 | 218 |

**Profile 4**

Rows: classification in generated profile; columns: classification in real profile

| | A | B | S |
|---|---|---|---|
| A | 866 | 59 | 0 |
| B | 24 | 28990 | 0 |
| S | 0 | 33 | 28 |

**Profile 5 (least intense)**

Rows: classification in generated profile; columns: classification in real profile

| | A | B |
|---|---|---|
| A | 565 | 75 |
| B | 13 | 29347 |

*Figure 8: Confusion matrix showing classification of scan points in a profile from its original form and generated form. Matrices are shown for 5 profiles, ranging from intense (first) to weak (last). Classification categories are A = allele, B = baseline, F = Forward stutter, H = half back stutter, P = pull-up and S = back stutter. The shading of the cells represent the number of observations from black (high) to white (low).*

The aim of creating the generator was to be able to generate realistic looking electropherograms with contributors that were simulated to have specific alleles. This final aspect was tested using the R package simDNAmixtures, which hold a series of biological models that dictate the DNA profile information that could be obtained from a series of contributors with specific genotypes. The models within simDNAmixtures include the generation of stutters and peak height imbalance that are expected in data produced within a laboratory. Figure 9 shows an example of DNA profile information generated using simDNAmixtures, which is then converted to a smoothed profile by drawing a normal distribution around the position of each simulated peak (smoothed profile not shown in Figure 9), and then passed into the generator to produce a realistic electropherogram. The generation of these profiles were successfully carried out, and there were many instances of baseline drift added into these profiles.

| Locus | Allele | Height | Size |
|---|---|---|---|
| D3S1358 | 14 | 250 | 117.33 |
| D3S1358 | 15 | 128 | 121.4 |
| D3S1358 | 16 | 215 | 125.48 |
| D3S1358 | 17 | 300 | 129.56 |
| D3S1358 | 18 | 55 | 133.64 |
| vWA | 16 | 89 | 176.91 |
| vWA | 17 | 111 | 180.95 |
| vWA | 18 | 139 | 184.99 |
| D16S539 | 10 | 121 | 247.64 |
| D16S539 | 11 | 248 | 251.67 |
| D16S539 | 12 | 135 | 255.7 |
| D16S539 | 13 | 112 | 259.73 |
| CSF1PO | 10 | 111 | 298.34 |
| CSF1PO | 11 | 487 | 302.3 |
| CSF1PO | 12 | 112 | 306.26 |
| TPOX | 8 | 940 | 349.7 |
| TPOX | 9 | 466 | 353.72 |
| Yindel | 2 | 534 | 86.79 |
| D8S1179 | 12 | 67 | 143.16 |
| D8S1179 | 13 | 188 | 147.26 |
| D8S1179 | 14 | 90 | 151.36 |
| D21S11 | 30 | 254 | 207.69 |
| D21S11 | 31.2 | 142 | 212.54 |
| D18S51 | 12 | 146 | 281.63 |
| D18S51 | 17 | 68 | 301.83 |
| DYS391 | 10 | 574 | 376.99 |
| D2S441 | 10 | 796 | 85.37 |
| D2S441 | 11 | 95 | 89.42 |
| D2S441 | 12 | 113 | 93.48 |
| D19S433 | 11 | 64 | 137.78 |
| D19S433 | 12 | 1274 | 141.77 |
| D19S433 | 13 | 335 | 145.75 |
| D19S433 | 14 | 182 | 149.74 |
| TH01 | 6 | 243 | 186.89 |
| TH01 | 9 | 98 | 199.38 |
| TH01 | 9.3 | 125 | 200.62 |
| FGA | 20 | 187 | 251.87 |
| FGA | 21 | 208 | 255.94 |
| FGA | 23 | 74 | 264.08 |
| D22S1045 | 11 | 146 | 97.51 |
| D22S1045 | 15 | 508 | 109.46 |
| D5S818 | 11 | 95 | 154.87 |
| D5S818 | 12 | 868 | 158.92 |
| D5S818 | 13 | 151 | 162.97 |
| D13S317 | 8 | 741 | 210.84 |
| D13S317 | 9 | 180 | 214.88 |
| D13S317 | 11 | 78 | 222.97 |
| D13S317 | 12 | 715 | 227.02 |
| D7S820 | 8 | 83 | 270.42 |
| D7S820 | 9 | 89 | 274.39 |
| D7S820 | 11 | 105 | 282.34 |
| D7S820 | 12 | 55 | 286.32 |
| SE33 | 27.2 | 80 | 400.19 |
| SE33 | 29.2 | 90 | 408.32 |
| D10S1248 | 13 | 322 | 105.53 |
| D10S1248 | 16 | 57 | 117.53 |
| D1S1656 | 11 | 94 | 168.02 |
| D1S1656 | 14 | 101 | 180.67 |
| D1S1656 | 15 | 189 | 184.88 |
| D1S1656 | 16 | 341 | 189.1 |
| D1S1656 | 18.3 | 74 | 198.8 |
| D12S391 | 19 | 343 | 236.04 |
| D12S391 | 20 | 1389 | 240.01 |
| D12S391 | 21 | 282 | 243.99 |
| D2S1338 | 20 | 50 | 316.84 |
| D2S1338 | 25 | 83 | 336.93 |
| AMEL | X | 306 | 106 |
| AMEL | Y | 227 | 113 |

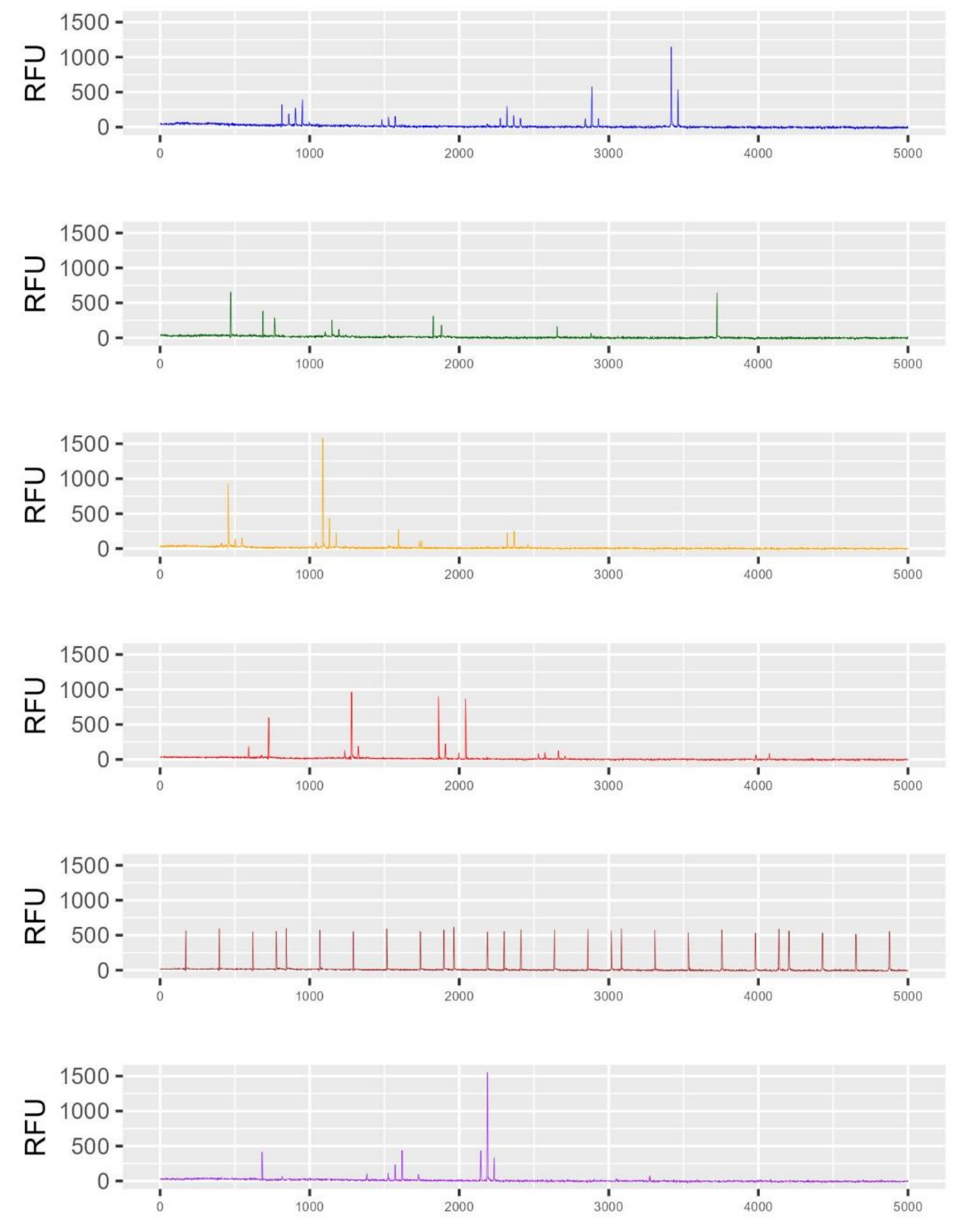

*Figure 9: DNA profile generated using simDNAmixtures showing the simulated peak information (left), and the final output using the generator (right)*

## 4.0 - Discussion

4.1 – performance of the GAN

The use of a GAN to create an ANN that can apply a ‘realism filter’ to a fake DNA profile data was ultimately successful. Earlier trials to train the same GAN system on 600 profiles consistently failed and given that the current system using 1078 profiles succeeded with no change other than the training dataset it can be surmised that the performance of the system is sensitive to the amount of training data. Even in the training dataset of 1078 there are several indications that further data is needed. One example is shown in Figure 8, where between 25% and 50% of pull-up features in the original profile were recreated in the generated profile.

Another indication that further training data may be required is that often, original profiles which exhibited baseline drift had generated profiles that also exhibited similar baseline drift. As there is no indication of baseline drift in the smoothed profiles, this suggests that the generator may be recognising the pattern of legitimate peaks in the input profiles as a means of determining if drift is added. This is a classic instance of overfitting to the dataset.

A third indication is that, while profile can be generated using simDNAmixtures and will have different patterns of baseline drift (as well as different patterns of baseline noise), if the same input profile is regenerated multiple times with different random number arrays for the secondary input, the pattern of baseline drift is always the same. This may be a result of the finding of the pix2pix authors that random inputs into the pix2pix GAN are effectively ignored, however the baseline noise component of profiles does change with the different random secondary inputs, and so the stable baseline drift may also simply be a product of too few training examples. Because the baseline noise component of the generated profiles varies with the random array of secondary input, it allows multiple attempts to be made at generating a profile with realistic baseline, using the discriminator ANN as a decision tool i.e. a set of peaks is simulated using simDNAmixtures and then the generator is used to generate 1000 versions of this profile as an electropherogram. With each generation the discriminator is used to classify it as real of fake and then the generated electropherogram with the highest discriminator output is the one chosen for any downstream use.

There is a possibility for the biological models within simDNAmixtures and the machine learning models in the generator to work against each other. For example, the biological models have a component of peak height variability that can lead to low-level peaks being absent from the profile (a phenomenon known as ‘drop-out’). In Figure 6 an instance was shown where a stutter peak that had dropped out in the real profile was added back into the generated profile as the generator had learnt that stutter peaks typically precede larger parent peaks. Again, this may be a problem that is addressed by increased training data, as eventually the discriminator will learn that if a profile has a missing stutter peak, then it must be real. This will then lead to the generator learning to sporadically dropout stutter peaks. This will mean that while the biological model drops out peaks sporadically, the machine learning models will sporadically add them back in and drop out others.

### 4.2 – further modification of the pix2pix GAN

In the pix2pix network the generator loss function is a combination of loss coming from the ability of the discriminator, and the closeness of the generator profile to the real image. In their original publication the authors of the pix2pix network (Isola et al., 2018) weighted the component relating to the alignment with the real image as 100 times higher than the component relating to the ability of the discriminator ANN. In our application to electrophoretic data we found that the weighting between the two components had to be equal. When the weighting was too far in favour or replicating the real profiles then the generator would generate profiles that had no randomised features (such as baseline noise and baseline drift) and would only learn the structured patterns in the data. Such a system could be taken further so that the discriminator is removed altogether and just the generator is trained in isolation (i.e. not in a GAN architecture). This may be a useful tool to learn the basic structural features of DNA profiles without the complication of random noise elements. It could also be used to generate foundational data that are used by other algorithms to generate the random or noise components, for example it may be that a time series could model baseline noise, or it may be that a GAN with a different structure to the pix2pix GAN is best suited for random noise generation elements (such the waveGAN (Donahue, McAuley, & Puckette, 2019), which has been used to generate different types of signal noise).

### 4.3 – The potential for other generative models

While our work focussed on the pix2pix GAN model architecture there are other models that could be trialled to generate EPG data. Many of these systems have the ability to generate realistic looking data but possess a reduced ability for generating data whose outcome can be carefully controlled in the same way that the pix2pix GAN. One potential alternative model that could be used is the CycleGAN (Zhu, Park, Isola, & Efros, 2017). CycleGAN does not require paired images, and instead uses the ‘style’ of groups of images within the two domains (in the context of our work this would be the idealised domain and the realistic domain). The model works by the first generator converting images from domain one to domain two, and the discriminator distinguishing between images natively belonging to domain two and generated fakes. The ability to retain features of the original image (from domain one) then comes from a second generator, which takes the fake image that has been generated to appear as though it belongs to domain two and converts it back to a domain one image. A second discriminator then distinguishes between the image belonging to domain one that has been converted to and from domain two and images natively belonging to domain one. CycleGAN is often chosen as a style translation tool due to the fact that it does not require paired data, which can be costly to produce. CycleGAN relies on the optimisation of the structure of the neural network to preserve the original image features, rather than relying on the pairing of the dataset as in pix2pix (Lin, 2023). As such pix2pix provides a better ability to control the fine scale features of the output than CycleGAN. This is the desired behaviour of the current application, where EPGs with specific peaks at specific heights are desired as the output.

Diffusion models represent another means of potentially generating EPG data and can produce more realistic synthetic data than GANs (Dhariwal & Nichol, 2021). Diffusion models work by adding sequentially more noise to an original image according to a schedule and training a network to identify the noise within one of the noised images. Using this network an input of random noise can then be 'de-noised' iteratively until a realistic image is obtained. The popularisation of diffusion models comes from the ability to condition their generation, typically using text prompts. For the application of generating EPG data the conditioning information could be provided in the form of the idealised, smoothed version of the profile with the diffusion model then producing the realistic synthesised data. Some optimisations would be required to balance the synthesis of realistic data and the retention of the conditioned features.

There also exist a number of GANs that have been specifically designed for generation of time series data such as TSGAN (Smith & Smith, 2020), GT-GAN (Jeon, Kim, Song, Cho, & Park, 2022), or TTS-CGAN (Li & Ngu, 2022), but again don't provide the same level of fine-scale control on the synthesised data that is provided by the pix2pix network.

4.4 – **Use cases** of the current GAN

As well as the use of the generator ANN to create training material for other ANNs, the profiles could be generated for other purposes. One such purpose would be for the ongoing training and proficiency testing of analysts within forensic laboratories. It is common for laboratories to regularly construct mixtures with a known number of contributors, with known genotypes, donating known DNA amounts and use these to train and test the proficiency of their analysts at assigning the number of contributors to profiles, and interpreting potential donor genotypes. Generating these training profiles is time-consuming and costly as each round of generation requires informed volunteer consent, laboratory work and use of costly reagents. Being able to construct profile electronically that are indistinguishable from real profiles by analysts therefore has great potential resource advantage. The generated information could be inserted into instrument files with the use of tools such as sequinR (Charif & Lobry, 2007). The profiles could then be provided to analysts in a completely blinded manner (i.e. the analysts do not know which profiles are those they are being test on), which has been shown to provide a better indication of realistic performance (Mejia, Cuellar, & Salyards, 2020).

As with all realistic generative capability there is the flipside; with the ability to create realistic electropherograms and embed them into instrument files for training, so too is it possible to create deep-fake profiles for ill intent. Like many generative tools it may be wise to build in tags for generated profiles that identify them as generated to prevent any nefarious use.

## 5.0 - Conclusion

The purpose of this work was to be able to generate profiles in order to be able to create unlimited pre-labelled electrophoretic data for use in training a DNA profile classification ANN, such as that in (Taylor, 2022). The profiles simulated using simDNAmixtures will have information that allows labels to be provided for stutter and allele, however not pull-up. Pull-up may still require labelling either manually, or by the current profile classification ANN. This latter option is quite practical as pull-up is well classified in the ANN trained in (Taylor, 2022), and the main training needed to improve performance is for the ANN to distinguish low-level peak data from baseline noise. This will allow the ANN to be used for electropherogram classification of peaks to sub-30-rfu bounds, which appears to be the current barrier to performance based on data in (Taylor & Buckleton, 2023) (see their Figure 7).

Further work planned is to apply explainable AI techniques to determine what features of the electrophoretic data the discriminator is using in the generated and real profiles in order to be able to discriminate between them (Linardatos, Papastefanopoulos, & Kotsiantis, 2021). This may provide insight into ways that the generator can be improved.

## Acknowledgements

Points of view in this document are those of the author and do not necessarily represent the official position or policies of their organisations. Thanks to Maarten Kruijver for assistance with the setup and use of simDNAmixtures. Thanks to internal organisational reviewers and external reviewers for their comments and suggestions that improved the quality of this work.